# A Tale of Two Lexica: Testing
# Computational Hypotheses with Deep Convolutional Neural Networks

**Enes Avcu**

Department of Neurology,
Massachusetts General Hospital

eavcu@mgh.harvard.edu

**Olivia Newman**

Department of Neurology,
Massachusetts General Hospital

onewman@mgh.harvard.edu

**David Gow**

Department of Neurology,
Massachusetts General Hospital

dgow@helix.mgh.harvard.edu

## Abstract

Gow's (2012) dual lexicon model suggests that the primary purpose of words is to mediate the mappings between acoustic-phonetic input and other forms of linguistic representation. Motivated by evidence from functional imaging, aphasia, and behavioral results, the model argues for the existence of two parallel wordform stores: the dorsal and ventral processing streams. In this paper, we tested the hypothesis that the complex, but systematic mapping between sound and articulation in the dorsal stream poses different computational pressures on feature sets than the more arbitrary mapping between sound and meaning. To test this hypothesis, we created two deep convolutional neural networks (CNNs). While the dorsal network was trained to identify individual spoken words, the ventral network was trained to map them onto semantic classes. We then extracted patterns of network activation from the penultimate level of each network and tested how well features generated by the network supported generalization to linguistic categorization associated with the dorsal versus ventral processing streams. Our preliminary results showed both models successfully learned their tasks. Secondary generalization testing showed the ventral CNN outperformed the dorsal CNN on a semantic task: concreteness classification, while the dorsal CNN outperformed the ventral CNN on articulation tasks: classification by onset phoneme class and syllable length. These results are consistent with the hypothesis that the divergent processing demands of the ventral and dorsal processing streams impose computational pressures for the development of multiple lexica.

**Keywords**: Lexical Interface, Articulation, Semantics, Parallel Lexica, Dorsal Stream, Ventral Stream, CNNs, Deep Learning.

## 1 Introduction

The dual-stream model of Hickok and Poeppel (2007) proposes that there are two major processing streams for spoken language: (i) the dorsal stream that mediates the bi-directional mapping between sound and articulation, and (ii) the ventral stream that maps the bi-directional relationship between sound and meaning. Within this framework, Hickok and Poeppel (2007) identify a lexical interface region located in the posterior middle temporal gyrus (pMTG) and adjacent cortices as a component of the ventral stream. This interface region plays a functional role akin to Elman's (2004) notion of the lexicon as a mechanism for linking input to meaning, rather than a static repository of meaning.

Gow's (2012) dual lexicon model extends the dual-stream models of language processing and synthesizes evidence from aphasia, behavioral and neural results to identify two parallel wordform areas. In the dorsal processing stream, the supramarginal gyrus (SMG) and inferior parietal lobe mediate the mapping between sound and word-level articulatory representation. Whereas, in the ventral processing stream, the posterior middle temporal gyrus (pMTG) mediates the mapping between sound and semantic or syntactic representation. In this paper, we ask why two separate wordform areas are necessary. Words provide a useful level of representation for organizing processing in both streams, but it is not clear why this function cannot be played by a single wordform area. The division between interpretation-based ventral processing streams and action-based dorsal processing streams appears to predate the evolution of language in both visual and auditory processing (Rauschecker & Scott, 2009; Sheth & Young, 2016). This raises the possibility that functional anatomical factors created a bias towards anatomical stream separation. In addition to these potential anatomical constraints, we hypothesize that computational constraints shaped the development of parallel wordform areas that rely on different featural representations of words to mediate different mappings between sound and higher-order linguistic representations.

The mapping between sound and articulation is very different from the mapping between sound and meaning. While complex, the mapping between sound and articulation is relatively systematic and temporally contiguous and is strongly organized around the gestural units of the segment or syllable. In contrast, the mapping between sound and syntactic or semantic information, though partially systematic at the level of productive morphology, is largely arbitrary and primarily dependent on identifying larger temporal units. Given these differences, we believe it is likely that these mappings would optimally depend on different featural representations of wordform.

Deep neural networks provide a useful tool for exploring these questions. Several recent studies have used convolutional neural networks (CNNs) originally developed for image processing (Le Cun et al., 1989; Gu et al., 2018) to explore optimal feature spaces for the classification of naturalistic inputs, and their implications for functional specificity in cortical processing (Kell et al., 2018; Kell & McDermott, 2019; Dobs et al., 2019).

In this paper, we use CNNs to test the hypothesis that the complex, but systematic mapping between sound and articulation in the dorsal stream poses different computational pressures on feature sets than the more arbitrary mapping between sound and meaning. Our strategy is to train CNNs on broad dorsal (articulatory) and ventral (semantic) tasks and then use featural representations of individual words discovered by these models to train new classifiers on novel dorsal vs ventral discriminations. We predict that features from CNNs trained on dorsal mappings should have an advantage for categorization related to articulation but not semantic/syntactic categorization, whereas features from CNNs trained on ventral mappings should have an advantage for semantic/syntactic categorization but not categorization related to articulation.

## 2 Method

More than a half-million labeled excerpted speech clips mixed with background noise were used as the training data. The same tokens were used to train both the ventral and dorsal CNN networks. We have created a publicly available map of code duplicates in GitHub repositories (https://github.com/enesavc/cnn_duallexicon).

### 2.1 Training Data

All speech samples were extracted from the multi-speaker Spoken Wikipedia Corpus (SWC) of reading speech (Baumann et al., 2019). To ensure that there were sufficient tokens to train the classifier, we restricted the word set to items that occurred at least 200 times in the corpus. We also restricted the set to include items that occurred no more than 450 times to eliminate frequency disparities that could bias classifier performance. Based on these criteria, we ended up with 178 words (see Appendix for the list of all words).

For each token of a target word, we then created ten two-second audio clips by randomly jittering the onset of the word within the two-second clip such that part of the word overlapped with the midpoint of the sample. This training data augmentation technique meant that a minimum of 2000 targets could be generated from 200 tokens of a given word. We generated more than half a million total two-second clips. To improve the generalization, we then mixed these two-second clips with one of the three different kinds of background noise: (i) auditory scenes, (ii) instrumental music, (iii) multi-speaker speech babble. Following Kell et. al. (2018), we used the 2013 IEEE AASP Challenge on Detection and Classification of Acoustic Scenes and Events corpus (Stowell et al., 2015) for auditory scenes and corpus of public domain audiobook recordings (https://librivox.org/) to create the multi-speaker speech babble. For the instrumental music, we used the Instrument Recognition in Musical Audio Signals (IRMAS) corpus (Bosch et al., 2012) which includes predominant instruments like cello, clarinet, flute, acoustic guitar, electric guitar, organ, piano, saxophone, trumpet, violin. As in Kell et. al. (2018), the two-second clips were mixed with these background noises with randomly assigned SNR levels (a Gaussian with a standard deviation of 2 dB SNR and a mean of 10 dB (for speech babble) or 7 dB (for auditory scenes and music).

### 2.2 Training Tasks

The dorsal CNN was trained to differentiate between the 178 spoken words. We chose this task to draw attention to whole word phonological properties without explicitly requiring sublexical segmentation into phonemes or syllables. The ventral task involved 10-way semantic discrimination. We categorized the 178 words into 10 semantic domains derived from Thompson et. al. (2020)'s semantic domain analysis (see Appendix for the list of semantic domains). For example, the target words *album* or *radio* were assigned to the semantic domain 'media/arts', whereas the target words *region* or *station* were assigned to 'location'. There was an average of 18 target words per domain (median=17, minimum=9, maximum=26). The semantic class of media/arts contained the highest number of examples (around one hundred thousand) and the semantic class of time contained the lowest number of examples (around forty thousand examples).

## 2.3 The Input to the Network

We used cochleagrams of each two-second clip as the input to the network following Kell et. al. (2018). A cochleagram is a spectrotemporal representation of an auditory signal designed to capture cochlear frequency decomposition. We used this representation to ground the computational problem that is presented to higher-level auditory and speech processing areas in the human brain. Cochleagrams were created using the code-shared by Feather et. al. (2019). Each two-second clip was passed through a bank of 203 bandpass filters resulting in a cochleagram representation of 203 x 400 (frequency x time). See Fig.1 for a schematic representation of training data preparation.

## 2.4 The Network Architecture

We used CNNs comprised of convolution, normalization, pooling, and fully connected layers (see Kell et. al., 2018 for the definitions of the operations of each layer). In addition, we used Gaussian Noise layers to apply zero-centered Gaussian noise (0.1) after each convolutional layer during training. Our CNN architecture consisted of 21 layers (see Appendix for the hyperparameters of each layer). The final layer provided softmax classification based on the features developed over the preceding layers. Both the dorsal and ventral networks were identical in terms of architectural parameters except for the softmax layer, which consisted of 10 nodes in the ventral model and 178 nodes in the dorsal mode. After the fully connected

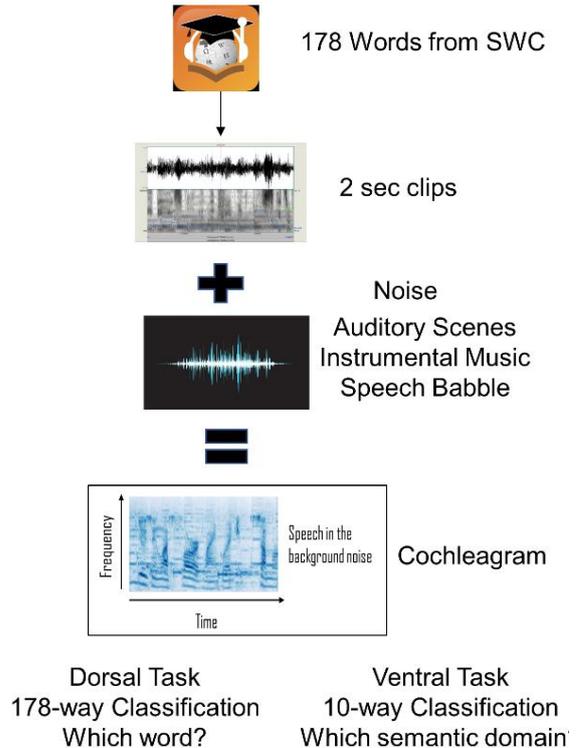

178 Words from SWC

2 sec clips

Noise
Auditory Scenes
Instrumental Music
Speech Babble

Cochleagram

Dorsal Task
178-way Classification
Which word?

Ventral Task
10-way Classification
Which semantic domain

**Figure 1**: Schematic representation of training data preparation and tasks for the networks.

layer, a dropout (0.1) layer was also used. Finally, we leveraged early stopping to avoid overfitting (when the validation loss does not decrease after ten consecutive epochs, the training stops and classifier weights from the epoch that has the lowest validation loss were saved).

## 2.5 Generalization Tasks

We aimed to compare the featural representations of both dorsal and ventral networks to test whether the networks learned similar or distinct features. To this end, we extracted the activation in the penultimate layer of each network (before the softmax layer where the classification happens) to secondary articulatory and semantic tasks. We then used a different set of softmax classifiers on the same featural representations to see whether the dorsal network learned distinct features related to articulation or the ventral network related to semantics.

For the articulatory generalization task, we used two tasks: (i) onset phoneme category (ii) syllable length. Onset phoneme category task was used to determine whether the first phoneme in a word was a fricative, nasal, stop, liquid, or vowel/glide (5 categories with 1050 exemplars each). A syllable length test was used to test whether words consisted of one, two, three, or four syllables (4 categories 900 exemplars each). For the semantic tasks, we use two tasks: (i) animacy and (ii) concreteness. The animacy task classified nouns as animate or inanimate (2 categories with 400 exemplars each). The concreteness task classified nouns as being abstract or concrete (2 categories with 500 exemplars each).

## 3   Results

### 3.1   CNNs Classification Accuracy

The results showed that both the dorsal and ventral networks had moderate validation accuracy levels (Fig.2). While the dorsal network was trained on a 178-way classification and had about 70% training and 50% validation accuracy, the ventral network was trained on a 10-way classification and had about 60% training and 18% validation accuracy. The reason for the ventral network's low validation accuracy compared to the dorsal network was due to the nature of the two tasks. Specifically, the identification of a word is a much easier task than the extraction of semantic domain information from different words. Nevertheless, both networks showed accuracy levels above chance.

### 3.2   CNNs Generalization Accuracy

Results of the generalization tasks showed that the features extracted from the dorsal network trained only on word recognition were more successful on articulatory tasks (dorsal CNN: 38% and 37% correct on onset class and syllable length tasks respectively) than ventral network (ventral CNN: 22% and 28% correct on onset class and syllable length tasks respectively). The features extracted from the ventral network trained only on semantic domains were more successful on the concreteness task (ventral CNN: 60% correct versus dorsal CNN: 39%). However, both networks showed comparable accuracy on the animacy task (ventral CNN: 64% correct versus dorsal CNN: 65%) which is not predicted (Fig.3).

These results suggest that the featural representations extracted from the dorsal network supported strong classification on secondary articulatory tasks compared to the ventral features. As for the secondary semantic tasks, ventral features were able to decode some novel semantic categories better than dorsal features.

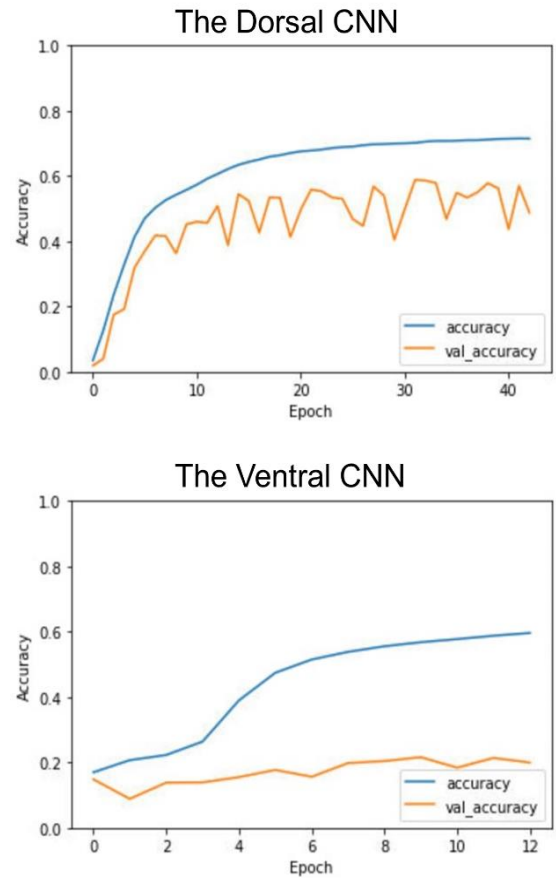

**Figure 2:** The classification accuracy of dorsal and ventral networks.

## 4   Discussion

In this study, we trained CNNs to map novel tokens of words in connected speech onto words and semantic classes. We tested the hypothesis that the complex, but systematic mapping between sound and articulation in the dorsal stream poses different computational pressures on feature sets than the more arbitrary mapping between sound and meaning. We first showed that identification of a word in a two-second clip mixed with noise is an easier task than extracting the semantic domain information in addition to word identification for a CNN. Next, we showed that the featural representation of words from CNNs trained on word recognition supported articulatory classification better than those trained on semantic categories. Conversely, featural representation of words from CNNs trained on semantic categories supported some semantic generalization better than those trained on word recognition. In general, these results are consistent with the claim that different featural projections of wordform may be needed to support efficient processing in the dorsal and ventral speech streams.

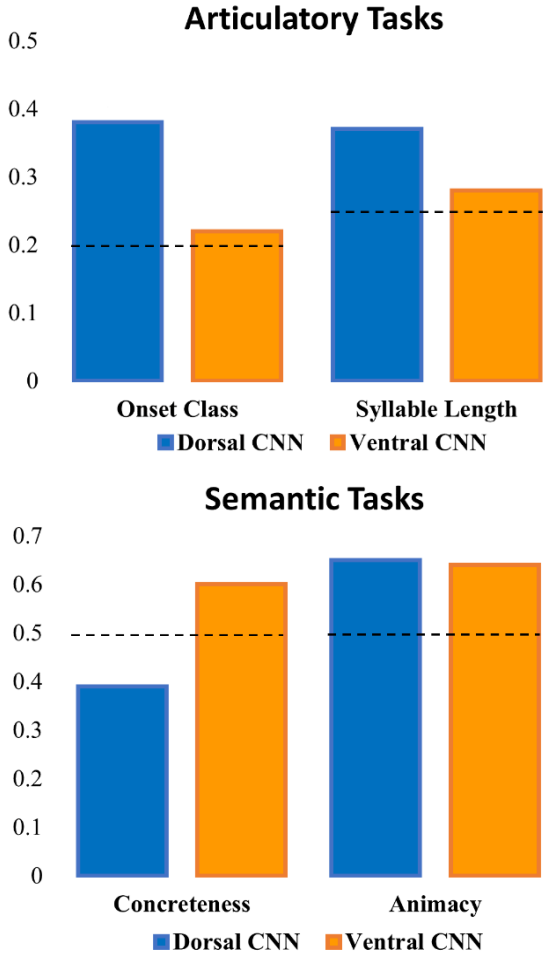

**Figure 3:** The generalization accuracy of dorsal and ventral networks. Classification accuracy based on features extracted from the penultimate layer of dorsal network is shown in blue. Accuracy based on ventral network features is shown in orange. The dashed black line indicates chance level in each secondary task.

In conclusion, our preliminary results are consistent with the proposal that the development of parallel lexica in the dorsal and ventral pathways arose from computational pressures for optimizing the primary mapping functions that support lexically organized processes in the dorsal and ventral processing streams.

## 5    Limitations and Future Directions

Although our preliminary results supported the hypothesis that different computational pressures play role in the emergence of two wordform areas, there are many limitations of this work that encourage future work.

At first, for the ventral CNN, we used a different classification task based on recognizing lexicosemantic representations derived from a distributional analysis of word co-occurrence (Mandera et al. 2017; Lenci, 2018). To this end, the ventral CNN network was trained to map individual words onto overlapping sets of collocates as a surrogate measure of semantic content. Items were classified based on co-occurrence with 438 collocate words identified in the billion-word Corpus of Contemporary American English (COCA) (Davies, 2020). Each word was trained for membership in 5 to 19 collocate categories that each overlapped across multiple words (nearly six and half million images). However, the model could not learn the semantic representations derived from a distributional analysis of word co-occurrence. We think the failure of this ventral CNN task reflects the complexity of using a large number of overlapping phonologically heterogeneous collocate categories. Meaningful comparison of ventral and dorsal training sets will require comparable individual word identification based on unique ventral and dorsal feature encodings. This will require improvements in training and network architecture to improve validation accuracy in both CNNs. Besides, expanding the focus on morphologically complex words to make sound meaning mappings more systematic and reflect natural language could produce a more direct model of speech processing.

The other limitation of the current study is the low accuracy of the ventral CNN model on the semantic domains task which is probably due to the lack of sufficient training data. We believe training the network on larger vocabulary (possibly requiring additional speech corpora) could yield better performance.

We also believe that replacing the CNN architecture with Long Short-Term Memory (LSTM) networks will yield better modeling. Our ongoing analyses are testing the use of LSTM models to predict the word based on parts of it as a dorsal task and predict the next word based on a window of previous and following words as a ventral task. Based on Elman (1991, 2004), we predict that word prediction training will support the development of representations that support both semantic and syntactic/lexical information in the ventral path.

Finally, following Kell et. al. (2018), we ultimately hope to compare human and classifier error patterns and use the classifiers to predict cortical responses in SMG and pMTG. We hypothesize that feature optimization creates pressure for the emergence of multiple lexica but believe that anatomical dissociations underlying stream segregation also contributes to the emergence of dual lexica.

## Acknowledgments

This work was supported by NIDCD grant R01DC 015455 and benefited from funding from NCRR grant P41RR14075. We would like to thank Alison Xin and Tim Giannatsis for their help in the data preparation step, and Alex Kell, Jenelle Feather, Ray Gonzales, and Arne Köhn for their invaluable advice.

## References

Baumann, T., Köhn, A. & Hennig, F. The Spoken Wikipedia Corpus collection: Harvesting, alignment and an application to hyperlistening. Lang Resources & Evaluation 53, 303 329 (2019). https://doi.org/10.1007/s10579-017-9410-y

Bosch, J. J., Janer, J., Fuhrmann, F., & Herrera, P. (2012). A Comparison of Sound Segregation Techniques for Predominant Instrument Recognition in Musical Audio Signals in Proc. ISMIR (pp. 559 564).

Davies, M. (2020). The Corpus of Contemporary American English. www.English corpora.org/coca/

Dobs, K., Kell, A., Palmer, I., Cohen, M., & Kanwisher, N. (2019). Why Are Face and Object Processing Segregated in the Human Brain? Testing Computational Hypotheses with Deep Convolutional Neural Networks. Oral presentation at Cognitive Computational Neuroscience Conference, Berlin, Germany.

Elman, J. L. (1991). Distributed representations, simple recurrent networks, and grammatical structure. Machine learning, 7(2), 195-225.

Elman, J. L. (2004). An alternative view of the mental lexicon. Trends in Cognitive Science, 8(7), 301-306. doi:10.1016/j.tics.2004.05.00

Feather, J., Durango, A., Gonzalez, R., & McDermott, J. (2019). Metamers of neural networks reveal divergence from human perceptual systems. In NeurIPS (pp. 10078-10089).

Gow, D. W. (2012). The cortical organization of lexical knowledge: A dual lexicon model of spoken language processing. Brain and language, 121 (3), 273 288. doi:10.1016/j.bandl.2012.03.005

Gu, J., Wang, Z., Kuen, J., Ma, L., Shahroudy, A., Shuai, B., ... & Chen, T. (2018). Recent advances in convolutional neural networks. Pattern Recognition, 77, 354-377.

Hickok, G. & Poeppel (2007). The cortical organization of speech processing. Nature Reviews Neuroscience, 8, 393 402. doi : 10.1038/nrn2113

Kell, A. J. E., Yamins, D. L. K., Shook, E. N., Norman-Haignere, S. V., & McDermott, J. H. (2018). A Task-Optimized Neural Network Replicates Human Auditory Behavior, Predicts Brain Responses, and Reveals a Cortical Processing Hierarchy. Neuron, 98(3), 630–644.e16.

Kell, A. J., & McDermott, J. H. (2019). Deep neural network models of sensory systems: windows onto the role of task constraints. Current Opinion in Neurobiology, 55, 121–132.

Le Cun, Y., Boser, B., Denker, J. S., Henderson, D., Howard, R. E., Hubbard, W., & Jackel, L. D. (1989). Handwritten digit recognition with a back-propagation network. In Proceedings of the 2nd International Conference on Neural Information Processing Systems (pp. 396-404).

Lenci, A. (2018). Distributional models of word meaning. Annual review of Linguistics, 4, 151 171.

Mandera, P., Keuleers , E., & Brysbaert , M. (2017). Explaining human performance in psycholinguistic tasks with models of semantic similarity based on prediction and counting: A review and empirical validation. Journal of Memory and Language, 92, 57 78.

Rauscchecker, J.P., & Scott, S.K. (2009). Maps and streams in the auditory cortex: Nonhuman primate illuminate human speech processing. Nature Neuroscience, 12(6), 718-724. doi: 10.1038/nn.2331

Sheth, B.R., & Young, R. (2016). Two visual pathways in primates based on sampling space: Expoitation and exploration of visual information. Frontiers in Integrative Neuroscience, 10(37). doi: 10.3389/fnint.2016.00037

Stowell, D., Giannoulis , E., Benetos, M., Lagrange, M., and Plumbley, D. (2015). Detection and Classification of Audio Scenes and Events. IEEE Transactions on Multimedia 17(10), 1733 1746, 2015.http://dx.doi.org/10.1109/TMM.2015.242899 8.

Thompson, B., Roberts, S. G., & Lupyan, G. (2020). Cultural influences on word meanings revealed

through large-scale semantic alignment. Nature Human Behaviour, 4(10), 1029-1038.

# A Appendices

## The List of Target Words Used for Network Training

above, according, against, album, among, another, between, century, certain, common, community, culture, developed, development different, during, early, election, episode, established, female, final, first, following, foreign, found, general, given, government, group, health, history, house, human, include, including, increase, independent, individual, influence, information, interest, international, issue, known, language, large, leader, leading, legal, limited, local, major, making, market, media, meeting, member, military, million, modern, music, national, natural, nature, nearly, north, northern, novel, nuclear, number, official, original, originally, other, outside, particular, party, performance, personal, playing, police, policy, political, popular, population, power, present, president, press, previous, primary, problem, process, production, program, project, proposed, provide, provided, public, published, radio, received, recent, record, recorded, region, related, release, released, religious, report, reported, required, research, result, school, science, scientific, series, service, several, significant, similar, small, social, society, software, sound, source, south, space, special, species, specific, stage, standard, state, stated, station, story, structure, study, subject, success, successful, support, system, technology, television, their, theory, these, third, three, throughout, title, total, trade, traditional, training, under, universe, variety, various, version, western, which, whose, widely, within, working, works, world, writing, written, young.

## Semantic Domains

*Academic*: according, found, given, novel, original, press, previous, program, proposed, provide, provided, published, received, related, report, reported, result, school, similar, special, specific, subject, success, successful, support, training. *Language*: language, required, source, stated, story, title, which, whose, writing, written. *Location*: above, against, among, between, north, northern, outside, region, south, space, station, under, universe, western, within, world. *Media/arts*: album, culture, episode, information, known, making, media, modern, music, originally, performance, playing, popular, production, radio, record, recorded, release, released, series, software, sound, stage, television, version, works. *Political*: election, government, independent, influence, issue, leader, legal, local, member, military, national, official, party, personal, police, policy, political, power, president, state, structure, system. *Community*: common, community, female, foreign, general, human, interest, market, meeting, population, problem, public, service, working. *Quantity*: another, certain, first, increase, large, limited, major, million, nearly, number, other, several, significant, small, these, third, three, total. *Science*: health, history, natural, nature, nuclear, project, research, science, scientific, species, study, technology, theory. *Social*: developed, development, different, established, group, house, include, including, individual, international, leading, particular, primary, process, religious, social, society, standard, their, trade, traditional, variety, various, widely. *Time*: century, during, early, final, following, present, recent, throughout, young.

## Network Architecture

- Input (203x400): Cochleagram: 203 frequency bins x 400 time bins
- Conv1 (68x134x96): Convolution of 96 kernels with a kernel size of 9 and a stride of 3
- Gaus1 (68x134x96): Gaussian noise (0.1)
- Norm1 (68x134x96): Normalization over 5 adjacent kernels
- Pool1 (34x67x96): Max pooling over window size of 3x3 and a stride of 2
- Conv2 (17x34x256): Convolution of 256 kernels with a kernel size of 5 and a stride of 2
- Gaus2 (17x34x256): Gaussian noise (0.1)
- Norm2 (17x34x256): Normalization over 5 adjacent kernels
- Pool2 (9x17x256): Max pooling over a window size of 3x3 and a stride of 2
- Conv3 (9x17x512): Convolution of 512 kernels with a kernel size of 3 and a stride of 1
- Gaus3 (9x17x512): Gaussian noise (0.1)

- Norm3 (9x17x512): Normalization over 5 adjacent kernels
- Pool3 (5x9x512): Max pooling over a window size of 3x3 and a stride of 2
- Conv4 (5x9x1024): Convolution of 1024 kernels with a kernel size of 3 and a stride of 1
- Gaus4 (5x9x1024): Gaussian noise (0.1)
- Norm4 (5x9x1024): Normalization over 5 adjacent kernels
- Conv5 (5x9x512): Convolution of 512 kernels with a kernel size of 3 and a stride of 1
- Gaus5 (5x9x512): Gaussian noise (0.1)
- Norm5 (5x9x512): Normalization over 5 adjacent kernels
- Pool4 (3x5x512): Mean pooling over a window size of 3 and a stride of 2
- Dense1 (4096): A fully connected layer
- Dense2 (178 or 10): A fully connected layer before the softmax function for words (n = 178) or semantic domains (n = 10).